\documentclass{llncs}

\usepackage{graphicx}

\usepackage{url}

\begin{document}

  \title{Reasoning for Moving Blocks Problem:\\ Formal Representation and Implementation}
  \titlerunning{Reasoning for Moving Blocks Problem}

  \author{PRZEMYS\L{}AW A. WA\L{}\c{E}GA}
  \authorrunning{P. A. Wa\l{}\c{e}ga}

  \institute{Institute of Philosophy, University of Warsaw, POLAND\\
  Institute of Automatic Control and Robotics. Warsaw University of Technology, POLAND}

\nocite{*}

\maketitle

  \begin{abstract}
  The combined approach of the Qualitative Reasoning and Probabilistic Functions for the knowledge representation is proposed. The method aims at represent uncertain, qualitative knowledge that is essential for the moving blocks task's execution. The attempt to formalize the commonsense knowledge is performed with the Situation Calculus language for reasoning and robot's beliefs representation. The method is implemented in the Prolog programming language and tested  for a specific simulated scenario. In most cases the implementation enables us to solve a given task, i.e., move blocks to desired positions. The example of robot's reasoning and main parts of the implemented program's code are presented.
  \end{abstract}


\section{Introduction}
\label{Introduction}

Reasoning algorithms that may represent incomplete and uncertain knowledge are greatly needful in a number of robotic applications, where the exact information is hardly available. In such cases, the probabilistic reasoning methods are mostly in use because of their ability to model uncertain knowledge and cognitive agent's beliefs. In parallel, there are also many attempts to develop qualitative approaches inspired by the human way of interpreting the surrounding world. Such methods allow to model imprecise and uncertain information, vary the grain of the reasoning, eliminate information overload, and fasten program execution (see \cite{Forbus1} and \cite{Forbus2} for details). The qualitative approach is still being explored, therefore a number of new methods are lately developed as presented in \cite{bhatt}, \cite{PDL}, \cite{jgp},  and \cite{Belief}, among others.

The proposed approach is a combination of both approaches: Qualitative Reasoning and Probabilistic Functions. The developed method for beliefs' representation is universal but was implemented and tested for a specific scenario in which the robot's task is to move cubic blocks in order to obtain the goal configuration using qualitative reasoning methods, with no numerical, quantitative information. One of the main advantages of the approach is the formal language that describes the robot's knowledge as a set of different, often inconsistent, beliefs. The formalism is based on the second order Situation Calculus language (\cite{Handbook}, \cite{Reiter}) that was established in order to reason about actions and change. The main concepts to be used are:
\begin{itemize}
  \item \emph{Actions} that may change the world features,
  \item \emph{Situations} which are possible states of the world that are in fact sequences of actions, called histories of the world,
  \item \emph{Fluents} that describe features of a specified situation.
\end{itemize}
\noindent In the Situation Calculus language the initial situation $S_{0}$ is an empty sequence of actions. The binary function symbol \texttt{do} maps a pair $(a,s)$, for an action $a$ and a situation $s$, to a successor situation obtained by performing $a$ to $s$. For example, the situation that occurs after performance of an action $a_1$ in the initial situation $S_{0}$, followed by $a_2$ and $a_1$ once more, is denoted by $\mathtt{do}(a_1, \mathtt{do}(a_2, \mathtt{do}(a_1, S_0)))$. Actions are usually parametrized, e.g., moving a block from a column $n$ to the top of a column $n'$ may be represented as an expression $\mathtt{move}(n,n')$, where $\mathtt{move}$ is a binary function symbol. The situation that results from moving block from a column $n$ to a column $n'$, and then moving it back from $n'$ to $n$ is denoted by $\mathtt{do}(\mathtt{move}(n',n),( \mathtt{do}(\mathtt{move}(n,n'), S_0)))$. Fluents may be functional or relational, but we will use only functional fluents that return values describing the world's features in a specified situation. For instance, the robot's belief that  a column $n$ is large in a situation $s$ can be represented as the functional fluent $\mathtt{large}(n,s)$ whose value describes a degree of the robot's belief. In our approach robot's beliefs about columns of blocks are described with qualities $\mathtt{Large}$, $\mathtt{Medium}$, $\mathtt{Small}$, and $\mathtt{Zero}$, although values of the corresponding functional fluents are expressed in terms of probability. Therefore, we develop a reasoning method that is based on the probabilistic approach. Furthermore, since the similarity to fuzzy logic was observed, the idea of triangular membership functions is used (see \cite{Fuzzy}). The proposed approach is not only developed but also implemented in the Prolog declarative programming language, which is often used to model artificial intelligence methods \cite{Bratko}. Therefore, it may be tested and the performance quality may be measured in different simulated situations. 

Most significant observed advantages of the proposed approach are: scalability feature, incomplete and uncertain knowledge representation, reasoning without strictly numerical information. Mentioned pros are essential for those robotic systems that need to work in the unknown environment or have information that is not enough precise to reason with classical quantitative methods. On the other hand, since the probabilistic reasoning algorithms are implemented in order to model the uncertainty, in some cases the robot possesses incorrect beliefs and is unable to complete the given task. Although the disadvantages cannot be ignored, the developed method has a number of significant features that are crucial for robotic application.

The paper is organized as follows. In Section \ref{Formal} the formal approach of representing beliefs, its language and axioms are introduced. Afterwards, solutions to the Qualification Problem and the Frame Problem are proposed. In Section \ref{Implementation} a brief breakdown of the Prolog implementation is shown. Three systems performance tests that present correct and faulty functionality are shown in Section \ref{Tests}. Finally, conclusions and future work plans are presented in Section \ref{Conclusions}. 

\section{The formal beliefs' representation in the Situation Calculus with probabilistic functions}
\label{Formal}

\begin{figure}[ht]
\centering
\includegraphics*[width=0.4\textwidth]{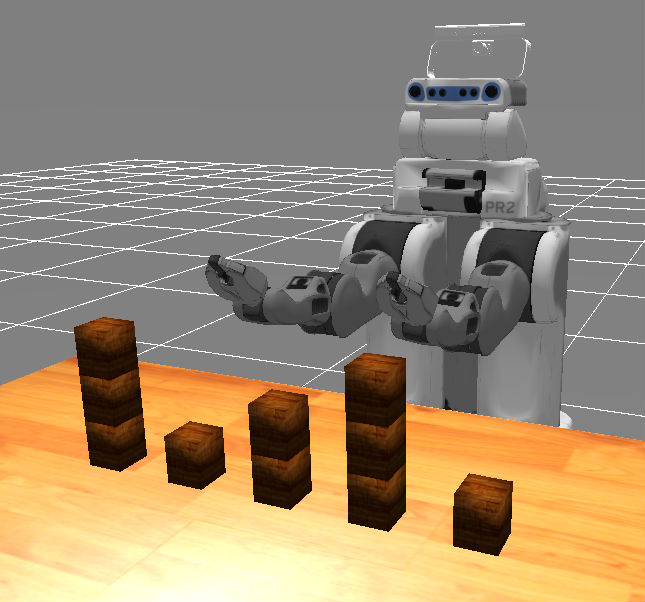}
\caption{The basic scenario setup.}
\label{fig::scenario}
\end{figure}

A specific scenario consists of a plane table with cubic blocks situated on its surface and the robot located in front of them. Separate columns are formed with blocks that are put on the top of each other as shown in Fig.~\ref{fig::scenario}.

\noindent The robot is equipped with an arm that can grab and move individual objects. At the beginning, blocks are placed on the table in the requested configuration which constitutes a goal setup. The robot observes the configuration and keeps its pure qualitative representation in mind. Afterwards, blocks are shuffled and a new configuration called an initial state is shown to the robot. The system's task is to observe the given configuration and plan the actions' sequence that will bring back the goal configuration i.e. in the final configuration height of each column will be the same as in the goal configuration. Predicates that are used for beliefs' description and reasoning procedure express qualitative (symbolic) information e.g. ''column 1 is large''. Therefore, the exact number of blocks in columns is unknown for the robot. Furthermore, since observations take place only twice -- when the goal and the initial states are presented -- the robot cannot check effects of performed actions.

The Situation Calculus formal approach is used in order to describe changes of robot's beliefs. The adapted second order language consists of the following alphabet:

\begin{itemize}
  \item Standard logical symbols: $\wedge$, $\neg$, $\exists$,
  \item Individual variables: $a$ for actions, $s$ for situations, $n$ for columns made of blocks,
  \item Constants $\mathtt{Large}$, $\mathtt{Medium}$, $\mathtt{Small}$, and $\mathtt{Zero}$ to represent qualitative values,
  \item Function symbols for situations: the initial situation $S_0$ and a binary function symbol $\mathtt{do}$, whose value $\mathtt{do}(a,s)$, for an action $a$ and a situation $s$, represents a situation $s'$ that results from applying an action $a$ in situation $s$,
  \item The binary predicate symbol $\subset$  for an ordering relation on situations (transitive, 
irreflexive, and asymmetric); $s \subset s'$ means that in  a situation $s$ performance of at least one action $a$ may lead to a situation $s'$; a binary predicate symbol $\subseteq$ is defined as usual: $s \subseteq s'$ iff $s = s'$ or $s \subset s'$,
  \item The binary predicate symbol $\mathtt{Poss}$; $\mathtt{Poss}(a,s)$ represents a possibility to perform an action $a$ in a situation $s$,
  \item The unary predicate symbol $\mathtt{col}$ for a situation independent relation; $\mathtt{col}(n)$ represents the fact that $n$ is a column,
  \item The binary function symbol $\mathtt{move}$ for a move action; $\mathtt{move}(n,n')$ means that the top block from a column $n$ is moved to the top of a column $n'$,
  \item Binary function symbols for functional fluents. Fluents describing agent's beliefs -- $\mathtt{large}$, $\mathtt{medium}$, $\mathtt{small}$, and $\mathtt{zero}$ -- map a pair $(n,s)$ to a value from the interval $\langle 0,1\rangle$ interpreted as a degree of robot's belief; $\mathtt{large}(2,s)=0.9$ means that in situation $s$ the robot has a strong belief (with a degree of 0.9) that a column 2 is $\mathtt{Large}$. Other fluents are used in order to describe the goal situation -- $\mathtt{goal}$ -- and main agent's beliefs -- $\mathtt{believe}$. Both of them are defined on pairs $(n,s)$ and return one of the qualitative values: $\mathtt{Large}$, $\mathtt{Medium}$, $\mathtt{Small}$, or $\mathtt{Zero}$, e.g. $\mathtt{goal}(2,s)=\mathtt{Large}$ means that in situation $s$ the robot's goal is to construct a column 2 that will $\mathtt{Large}$ and $\mathtt{believe}(2,s)=\mathtt{Small}$ means that in situation $s$ the robot's main believe about a column 2 is that it's $\mathtt{Small}$.
\end{itemize}

\begin{figure}[ht]
\centering
\includegraphics[trim = 0mm 203mm 75mm 0mm, clip, width=1\textwidth]{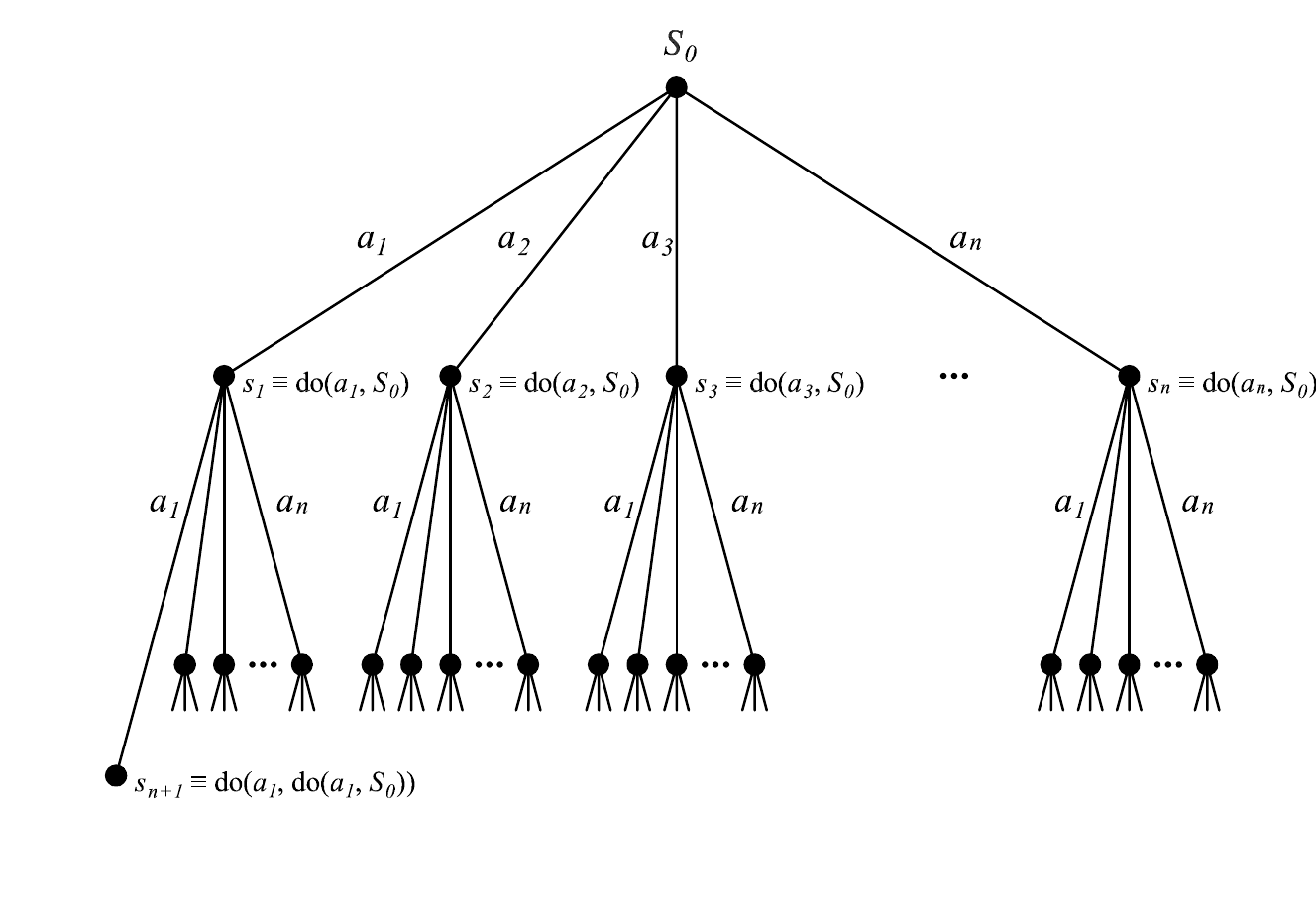}
\caption{A tree of situations graph.}
\label{fig::tree}
\end{figure}

\noindent The initial situation $S_0$ denotes the robot's beliefs and goals that were given to him at the beginning of the scenario. Performing different actions (that are in fact $\mathtt{move}$-actions done by the robot) leads to different situations, namely different agent's belief states. A~demonstrative tree of situations graph is presented in Fig.~\ref{fig::tree}.

The robot's task is to find a sequence of actions that leads to a situation in which his beliefs are equal to the given goals, thus to a situation in which $\mathtt{believe}(n) = \mathtt{goal}(n)$ holds for all $n$. If it is impossible to find such a path, then the robot's solution should bring him to the closest situation to the given goal. Fundamental axioms of the approach are shown in Fig.~\ref{fig::axioms}.

\begin{figure}[ht]
For all actions $a$ and situations $s,s', s_1, s_2, s_3$:

\medskip
\begin{tabular}{ll}
$S_0 \neq \mathtt{do}(a,s)$ &       \\
$\mathtt{do}(a,s)  \neq s$   &   \\
$s = S_0 \vee \exists a \exists s' (s = \mathtt{do}(a,s'))$ & (existence of a predecessor)\\
$S_0 \subseteq s$ & \\
$(s_1 \subset s_2 \wedge s_2 \subset s_3) \rightarrow  (s_1 \subset s_3)$ & (transitivity of $\subset$) \\
$\neg (s \subset s)$ & (irreflexivity of $\subset$) \\
$(s \subset s') \rightarrow \neg (s' \subset  s)$ & (asymmetry of $\subset$) \\
$(s \subseteq s') \leftrightarrow (s \subset s' \vee s=s')$ & \\
$\neg (\mathtt{do}(a,s) \subseteq s)$ & \\
$(s \subseteq s' \wedge s' \subseteq s) \rightarrow (s = s')$ & 
\end{tabular}
\caption{Fundamental axioms of the Situation Calculus.}\label{fig::axioms}
\end{figure}

The main idea of our method is to represent agent's beliefs in terms of the combined means: the qualitative and probabilistic (quasi-fuzzy) ones. Therefore, specific functional fluents for robot's beliefs are implemented. They return floating point numbers from the interval $\langle 0,1\rangle$ that are treated as a degree of the belief, where $0$ means the weakest belief and $1$ the strongest one. In the case of inconsistent beliefs occurrence -- such as ''column $n$ is large'' and ''column $n$ is small'' with the same degree of belief -- an additional, main agent's belief fluent $\mathtt{believe}$ with an argument $n$ is implemented. It may return one of the qualitative values: $\mathtt{Large}$, $\mathtt{Medium}$, $\mathtt{Small}$, or $\mathtt{Zero}$.

\begin{figure}[ht]
\centering
\includegraphics[trim = 0mm 220mm 95mm 10mm, clip, width=0.75\textwidth]{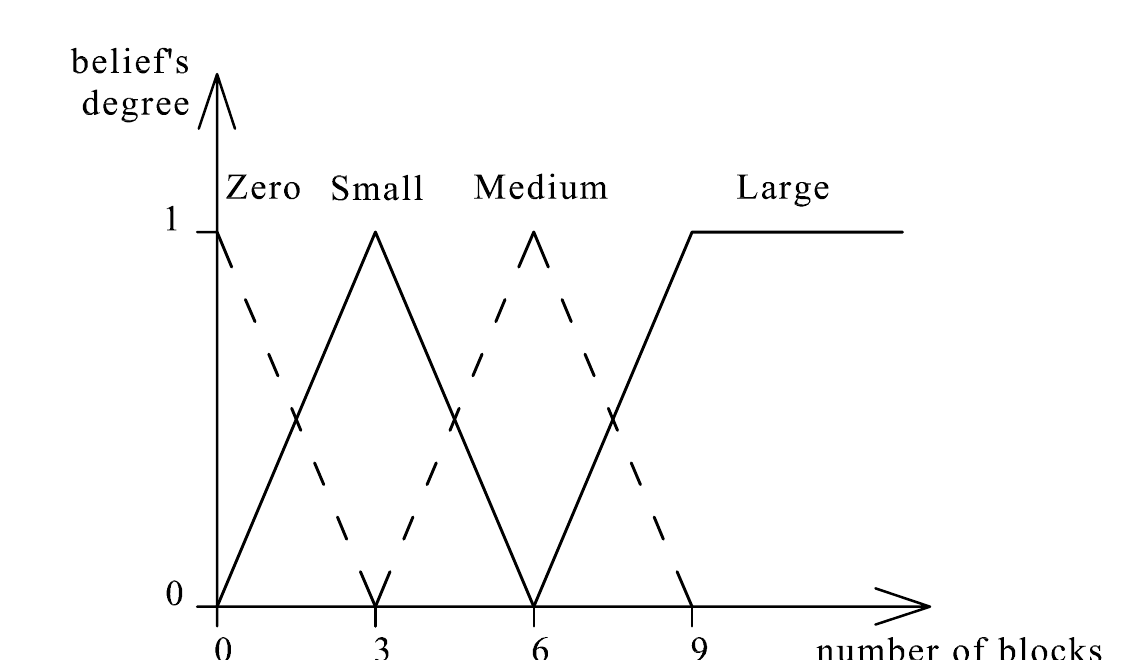}
\caption{Triangular membership functions for states of beliefs.}
\label{fig::triangular}
\end{figure}

\noindent The most desired situation occurs when values of belief fluents change triangular membership functions known from the fuzzy logic (see \cite{Fuzzy}), i.e., the increase of some belief function value leads to the reduction of  the ''nearest belief function'' value, where the ''nearest belief function'' corresponds to the next most probable state to occur. In Fig.~\ref{fig::triangular} we present an example, where the number of blocks in a column $n$ affects on agent's beliefs, namely $\mathtt{large}(n)$, $\mathtt{medium}(n)$, $\mathtt{small}(n)$, $\mathtt{zero}(n)$. Thus, adding the 9th block to the column increases a belief that this column is large and decreases a belief that it is medium.

Researches on the qualitative modelling and reasoning indicate that a pure qualitative approach is usually impractical and impossible to be implemented. Programmable methods used in the Qualitative Reasoning usually make use of a quantitative approach at least on some basic level. Although such implementations are more natural and intuitive due to usage of qualitative methods, numerical values still cannot be omitted and, as a consequence, hybrid approaches are most applicable. The proposed approach should be treated as such a hybrid method, since it combines qualitative and quantitative methods. Belief fluents refer to qualities such as ''column $n$ is large'', ''column $n$ is small'', whereas values of functional fluents are expressed with the quantitative probabilistic description. An action performance results with the belief fluents value changes which are precisely and numerically defined. 

\noindent One of the most desired feature of qualitative systems is the scalability (see \cite{Forbus1} for details). Not only the quantization needs to be relevant to the problem but also the ability of changing the grain of the reasoning should be provided. The fulfillment of such a constraint allows to change the modelling precision and is called \emph{scalability}. The proposed method provides such a feature, as the number of qualities and a degree of actions' influence on numerical fluents' values may be freely changed. The more qualities are, the more precise modelling becomes. 

There are three fundamental problems in the reasoning about actions: the \emph{Qualification Problem}, the \emph{Frame Problem} and the \emph{Ramification Problem}. In order to exploit the Situation Calculus formal language all of them need to be considered and solutions for them ought to be proposed.

The Qualification Problem states that it is impossible to define all preconditions of an action. As a result, no inference of the form $\mathtt{Poss}(a,s)$ about executable actions $a$ in a particular situation $s$ may be obtained. In order to explain the problem clearly and formally, let us consider an example in which the system contains the axiom (\ref{eqn:qp1}) of the form:\begin{equation}
\mathtt{Poss}(\mathtt{move}(n,n'),s) \rightarrow  (\mathtt{col}(n) \wedge \mathtt{col}(n') \wedge \neg (\mathtt{believe}(n,s)=\mathtt{Zero}))
    \label{eqn:qp1}
\end{equation} It means that if it is possible for the robot to move a block from a column $n$ to a column $n'$ in a situation $s$, then $n$ and $n'$ are columns and the robot believes that there are at least some blocks on the column $n$. Unfortunately, the reverse implication that is:\begin{equation}
(\mathtt{col}(n) \wedge \mathtt{col}(n') \wedge \neg (\mathtt{believe}(n,s)=\mathtt{Zero})) \rightarrow \mathtt{Poss}(\mathtt{move}(n,n'),s)
    \label{eqn:qp2}
\end{equation} is false. Hence, it is impossible to infer when a move action is executable. The condition~(\ref{eqn:qp2}) is unacceptable, since it may be impossible to perform a move action, for instance, when the top block on a column $n$ may be glued to the table or too heavy to be picked up by the robot. The number of such qualifications in a complex domain is extremely large, so it is obviously impossible to specify all of them. The preferable solution to the problem is to specify only important preconditions which are directly connected to an action's execution and ignore minor ones (until they arise). In most cases defining all preconditions is impractical or even impossible, whereas the presented solution allows to overcome both implementation and logical problems.

The Frame Problem concerns unchanged fluents after action's performance. A great number of such frame axioms need to be declared (their number is equal to the number of actions multiplied by the number of fluents). For instance, the axiom~(\ref{eqn:fp}) states that moving one block from a column $n$ to a column $n'$ does not affect the agent's main belief about the other column $n''$. \begin{equation}
\mathtt{believe}(n'',s) = \mathtt{believe}(n'', \mathtt{do}(\mathtt{move}(n,n'),s) ) 
    \label{eqn:fp}
\end{equation} \noindent The solution that was chosen is to assume that fluents may change only due to defined actions and no unknown effect may occur. That kind of solution is the simplest one, nevertheless it  is good enough for the developed approach.

The Qualification Problem is about indirect effects that are caused by actions performance e.g. moving the bottom block from a particular column leads to change of upper blocks position. A number of such effects cannot be provided, and may occur in some specific situations. Defining every single indirect effect is unreasonable and usually impossible. Therefore, our solution is to define a list of causal laws that describe indirect effects that occur when specific conditions are fulfilled. Summing up, direct effect are specified and additional causal laws for computing indirect effect are provided.
\section{The implementation}
\label{Implementation}

\noindent The implementation of the described scenario is done with the Prolog environment. Additionally, the IndiGolog \cite{Indigolog} -- an agent architecture with the Situation Calculus implementation, completely programmed in Prolog -- is used. It provides all elements of the Situation Calculus language and common programming structures like ''if'' condition and the ''while'' loop. The whole reasoning module is fully implemented in the IndiGolog architecture. Now, we will present and describe its parts. 

\medskip
\noindent \emph{Constants.} The only constants used in the program are columns numbers. The example presented in Fig.~\ref{constants} corresponds to the blocks' configuration shown in Fig.~\ref{fig::scenario}. It consists of five columns, where the expression $\mathtt{col}(\mathtt{N})$ means that $\mathtt{N}$ is a column.

\begin{figure}[ht]
\begin{verbatim}
  col(N)     :-    N=1; N=2; N=3; N=4; N=5.	
\end{verbatim}
\caption{Constants represent numbers of columns.}\label{constants}
\end{figure}

\noindent
\emph{Actions.} There is a single group of actions that may be performed by the robot for instance, moving the top block from a column $\mathtt{X}$ and putting it on the top of a column $\mathtt{Y}$. The Prolog definition is presented in Fig.~\ref{actions}.

\begin{figure}[ht]
\begin{verbatim}
  prim_action(move(X,Y))     :-    col(X),col(Y).	
\end{verbatim}
\caption{Action $\mathtt{move}(\mathtt{X},\mathtt{Y})$.}\label{actions}
\end{figure}

\noindent
\emph{Fluents.} One of the functional fluents describes the goal configuration to be achieved. It may return any quality available for a specified column. Other functional fluents denote beliefs' states of the robotic agent. The fluent $\mathtt{large}(\mathtt{N})$ returns a degree of a belief that a column $\mathtt{N}$ is large in a current state and it remains in the interval $\langle 0,1\rangle$ as described in Section~\ref{Formal}. The program code depicted in Fig.~\ref{fluents} shows additional three fluents: $\mathtt{medium}(\mathtt{N})$, $\mathtt{small}(\mathtt{N})$, and $\mathtt{zero}(\mathtt{N})$. They play the analogous role as $\mathtt{large}(\mathtt{N})$. One more functional fluent -- $\mathtt{believe}(\mathtt{N})$ -- is used to denote the most probable state of column $\mathtt{N}$ according to the robotic agent's knowledge. No relational fluents have been used.

\begin{figure}[ht]
\begin{verbatim}
  prim_fluent(large(N))     :-    col(N).	
  prim_fluent(medium(N))    :-    col(N).	
  prim_fluent(small(N))     :-    col(N).	
  prim_fluent(zero(N))      :-    col(N).		
  prim_fluent(goal(N))      :-    col(N).	
  prim_fluent(believe(N))   :-    col(N).	
\end{verbatim}
\caption{Functional fluents.}\label{fluents}
\end{figure}

\noindent
\emph{Causal laws.} The actions' influence on fluents' values has been described by making use of fixed laws. In fact, since  values of fluents denote agent's beliefs, causal laws expose the actions' influence on the robot's beliefs. Every action $\mathtt{move}(\mathtt{X},\mathtt{Y})$ changes robot's beliefs about the height of columns $\mathtt{X}$ and $\mathtt{Y}$, namely, column $\mathtt{X}$ is believed to decrease and column $\mathtt{X}$ to increase. Depending on earlier beliefs about $\mathtt{X}$ and $\mathtt{Y}$ height, different causal laws are to occur.

A short part of causal laws section is presented in the program code in Fig.~\ref{causal}. It describes how agent's beliefs change when a block from a column $\mathtt{X}$ is taken, if $\mathtt{X}$ was believed to be large so far. The probabilistic value of the functional fluent $\mathtt{large}(\mathtt{X})$ decreases and the value of the fluent $\mathtt{medium}(\mathtt{X})$ increases. Furthermore, if the value of $\mathtt{medium}(\mathtt{X})$ becomes greater than $0.5$ (or equivalently, greater than the value of fluent $\mathtt{large}(\mathtt{X})$), then the agent will believe that the column $\mathtt{X}$ is medium rather than large.

\begin{figure}[ht]
\begin{verbatim}
  causal_val(move(X,_),large(X),N,and(N is large(X)-0.25,large(X)>0.0)).	
  causal_val(move(X,_),medium(X),N,and(N is medium(X)+0.25,large(X)>0.0)).	
  causal_val(move(X,_),believe(X),medium,and((medium(X)+0.25)>0.5,large(X)>0.0)).	
\end{verbatim}
\caption{Examples of causal laws.}\label{causal}
\end{figure}

\noindent
\emph{Preconditions.} The action $\mathtt{move}(\mathtt{X},\mathtt{Y})$ is executable when a column $\mathtt{X}$ is not empty (see Fig.~\ref{poss}). It is worth noting that this precondition is constituted with respect to the robot's knowledge. Therefore, if the agent's beliefs are incorrect, he may try to take a block from an empty column.  

\begin{figure}[ht]
\begin{verbatim}
  poss(move(X,_))     :-    neg(believe(X)=zero) ).	
\end{verbatim}
\caption{An example of a precondition.}\label{poss}
\end{figure}

\noindent Belief functional fluents and the change of their values play the most significant role in our approach. Now, we will provide an illustrative example and its detailed description. Table~\ref{tab::beliefs} presents agent's qualitative belief states with respect to the precise number of blocks in a column $n$. As we mentioned in Section~\ref{Formal}, the proposed approach uses a scalability feature. In our case there are four qualities, namely $\mathtt{Large}$, $\mathtt{Medium}$,  $\mathtt{Small}$, and  $\mathtt{Zero}$. $\mathtt{Large}$ denotes any column whose number of blocks is in the interval $\langle 9,12\rangle$. Similarly, $\mathtt{Medium}$,  $\mathtt{Small}$, and  $\mathtt{Zero}$ denote any column whose number of blocks is in the interval $\langle 5,8\rangle$, $\langle 1,4\rangle$ and zero, respectively (see the top two ranks of Table~\ref{tab::beliefs}). There are four main agent's belief fluents related to a column $n$, i.e., $\mathtt{large}(n)$, $\mathtt{medium}(n)$,  $\mathtt{small}(n)$, and  $\mathtt{zero}(n)$. Thus, if in the initial situation shown to the agent the column $n$ has 11 blocks and he recognizes the column as large, then his initial beliefs should be as follows: $\mathtt{large}(n)=1$, $\mathtt{medium}(n)=0$,  $\mathtt{small}(n)=0$, and  $\mathtt{zero}(n)=0$ (see the third column of Table~\ref{tab::beliefs}). Performing an action $\mathtt{move}(n,n')$, i.e., taking the top block from a column $n$ and placing it somewhere else, results in a change of belief fluents' values. The magnitude of such a change depends on the granularity of qualities and equals $\frac{1}{\mathrm{granularity}}$. In our case the granularity equals $4$, so the magnitude is $0.25$. If one block from a column $n$ is moved, then $\mathtt{large}(n)$ decreases, while $\mathtt{medium}(n)$ increases. Thus, the final agent's beliefs should be: $\mathtt{large}(n)=0.75$, $\mathtt{medium}(n)=0.25$, $\mathtt{small}(n)=0$, $\mathtt{zero}(n)=0$ (see the fourth column of Table~\ref{tab::beliefs}). In a similar way, further $\mathtt{move}$-actions change the agent's beliefs as illustrated in Table~\ref{tab::beliefs}.

\begin{table}[ht]
\caption{Belief fluents table for 11 blocks in the initial state.}
\label{tab::beliefs}
\begin{minipage}{\textwidth}
\begin{tabular*}{1\textwidth}{@{\extracolsep{\fill} }  l  *{13}{c} }
\hline\hline
Qualities& \multicolumn{4}{|c|}{$\mathtt{Large}$}& \multicolumn{4}{c|}{$\mathtt{Medium}$}&  \multicolumn{4}{c|}{$\mathtt{Small}$}& $\mathtt{Zero}$\\
\hline
Blocks in col.& 12& 11& 10& 9& 8& 7& 6& 5& 4& 3& 2& 1& 0\\
\hline
$\mathtt{large}(n,s)$& & 1& 0.75& 0.5& 0.25& & & & & & & &  \\
$\mathtt{medium}(n,s)$& &  & 0.25& 0.5& 0.75& 1& 0.75& 0.5& 0.25& & & &  \\
$\mathtt{small}(n,s)$& & & & & & & 0.25& 0.5& 0.75& 1& 0.75& 0.5& 0.25  \\
$\mathtt{zero}(n,s)$& & & & & & & & & & & 0.25& 0.5& 0.75 \\
\hline\hline
\end{tabular*}
\end{minipage}
\end{table}

\noindent  The initial blocks setup has a significant meaning for the whole reasoning process. Table~\ref{tab::allbeliefs} shows values of belief fluents for four different initial setups, namely with 12 blocks, 11 blocks, 10 blocks, and 9 blocks in $n$. Although in all these cases the robot qualifies $n$ as large, further inferences about its height are different, and even in some cases reasoning leads  to false beliefs.

\begin{table}[ht]
\caption{Belief fluents table for four different initial states.}
\label{tab::allbeliefs}
\begin{minipage}{\textwidth}
\begin{tabular*}{1\textwidth}{@{\extracolsep{\fill} }  l  *{13}{c} }
\hline\hline
Qualities& \multicolumn{4}{|c|}{$\mathtt{Large}$}& \multicolumn{4}{c|}{$\mathtt{Medium}$}&  \multicolumn{4}{c|}{$\mathtt{Small}$}& $\mathtt{Zero}$\\
\hline
Blocks in col.& 12& 11& 10& 9& 8& 7& 6& 5& 4& 3& 2& 1& 0\\
\hline
$\mathtt{large}(n,s)$& 1& 0.75& 0.5& 0.25& & & & & & & & & \\
$\mathtt{medium}(n,s)$&  & 0.25& 0.5& 0.75& 1& 0.75& 0.5& 0.25& & & & & \\
$\mathtt{small}(n,s)$& & & & & & 0.25& 0.5& 0.75& 1& 0.75& 0.5& 0.25& \\
$\mathtt{zero}(n,s)$& & & & & & & & & & 0.25& 0.5& 0.75& 1\\
\hline
$\mathtt{large}(n,s)$& & 1& 0.75& 0.5& 0.25& & & & & & & &  \\
$\mathtt{medium}(n,s)$& &  & 0.25& 0.5& 0.75& 1& 0.75& 0.5& 0.25& & & &  \\
$\mathtt{small}(n,s)$& & & & & & & 0.25& 0.5& 0.75& 1& 0.75& 0.5& 0.25  \\
$\mathtt{zero}(n,s)$& & & & & & & & & & & 0.25& 0.5& 0.75 \\
\hline
$\mathtt{large}(n,s)$& & & 1& 0.75& 0.5& 0.25& & & & & & &  \\
$\mathtt{medium}(n,s)$& & &  & 0.25& 0.5& 0.75& 1& 0.75& 0.5& 0.25& & &   \\
$\mathtt{small}(n,s)$& & & & & & & & 0.25& 0.5& 0.75& 1& 0.75& 0.5  \\
$\mathtt{zero}(n,s)$& & & & & & & & & & & & 0.25& 0.5 \\
\hline
$\mathtt{large}(n,s)$& & & & 1& 0.75& 0.5& 0.25& & & & & & \\
$\mathtt{medium}(n,s)$& & & &  & 0.25& 0.5& 0.75& 1& 0.75& 0.5& 0.25& & \\
$\mathtt{small}(n,s)$& & & & & & & & & 0.25& 0.5& 0.75& 1& 0.75 \\
$\mathtt{zero}(n,s)$& & & & & & & & & & & & & 0.25\\
\hline\hline
\end{tabular*}
\end{minipage}
\end{table}

\section{Performance tests}
\label{Tests}

A number of tests for different initial and goal configurations were performed. The same scenario as a described one in Section~\ref{Formal} was simulated, i.e, there were five columns made of blocks and four qualities, namely $\mathtt{Large}$, $\mathtt{Medium}$,  $\mathtt{Small}$, and  $\mathtt{Zero}$. When the initial situation is shown to the robot, he assigns a quality value to each column and keeps in mind only such a qualitative representation. Therefore, the first column with 8 blocks in one initial situation and with 5 blocks in another initial situation is represented by the robot in both these cases in the same way, namely as a $\mathtt{Medium}$ column. This kind of system's behaviour may lead to mistaken reasoning and inability to complete a given task. Now, we present three different simulations, namely a goal achievement in a well established initial situation, a goal achievement in a borderline initial situation, and an inability to complete a given goal in a specific borderline initial situation.

Table~\ref{tab::goal1} shows a goal achievement simulation. Initial situation is well established for every column, which means that no borderline values are to be used. Borderline numbers of blocks in a $\mathtt{Medium}$ column are 5 and 8, since $\mathtt{Medium}$ denotes any column whose number of blocks is in the interval $\langle 5,8\rangle$. System's behaviour presented in Table~\ref{tab::goal1} leads to the goal achievement for every column. Another goal achievement simulation is presented in Table~\ref{tab::goal2}. Borderline numbers of blocks are established in order to make the goal achievement more difficult. Nevertheless, the given task is completed and no mistakes in blocks configuration for every column occur. At last in Table~\ref{tab::nogoal} a simulation, in which the goal is not achieved is presented. The robot obtains a goal for columns 2, 3, 4, and 5, but failed to place blocks in the first column properly. He believes that the final number of blocks in the first column that consists of 8 blocks corresponds to the $\mathtt{Large}$ quality but in fact it corresponds to the $\mathtt{Medium}$ quality. The robot's belief is incorrect, and therefore, the goal is not achieved. However, occasional mistakes cannot be eliminated as only the qualitative representation and not a quantitative exact number of block in each column is known to the robot.

\begin{table}[ht]
\caption{Goal achievement in a well established initial situation.}
\label{tab::goal1}
\begin{minipage}{\textwidth}
\begin{tabular*}{1\textwidth}{@{\extracolsep{\fill} }  l  *{5}{c} }
\hline\hline
Columns& 1& 2& 3& 4& 5 \\
\hline
Initially blocks in col.& 7& 2& 0& 11& 6 \\
Assigned qualities in initial sit.& $\mathtt{Medium}$& $\mathtt{Small}$& $\mathtt{Zero}$& $\mathtt{Large}$& $\mathtt{Medium}$ \\
Goal& $\mathtt{Small}$& $\mathtt{Small}$& $\mathtt{Medium}$& $\mathtt{Medium}$& $\mathtt{Small}$ \\
Finally blocks in col.& 4& 3& 8& 8& 3 \\
Goal achievement& yes& yes& yes& yes& yes \\
\hline\hline
\end{tabular*}
\end{minipage}
\end{table}

\begin{table}[ht]
\caption{Goal achievement in a borderline initial situation.}
\label{tab::goal2}
\begin{minipage}{\textwidth}
\begin{tabular*}{1\textwidth}{@{\extracolsep{\fill} }  l  *{5}{c} }
\hline\hline
Columns& 1& 2& 3& 4& 5 \\
\hline
Initially blocks in col.& 12& 8& 4& 9& 0 \\
Assigned qualities in initial sit.& $\mathtt{Large}$& $\mathtt{Medium}$& $\mathtt{Small}$& $\mathtt{Large}$& $\mathtt{Zero}$ \\
Goal& $\mathtt{Medium}$& $\mathtt{Medium}$& $\mathtt{Large}$& $\mathtt{Medium}$& $\mathtt{Small}$ \\
Finally blocks in col.& 7& 7& 11& 5& 3 \\
Goal achievement& yes& yes& yes& yes& yes \\
\hline\hline
\end{tabular*}
\end{minipage}
\end{table}

\begin{table}[ht]
\caption{Inability to complete a given goal.}
\label{tab::nogoal}
\begin{minipage}{\textwidth}
\begin{tabular*}{1\textwidth}{@{\extracolsep{\fill} }  l  *{5}{c} }
\hline\hline
Columns& 1& 2& 3& 4& 5 \\
\hline
Initially blocks in col.& 5& 3& 0& 7& 10 \\
Assigned qualities in initial sit.& $\mathtt{Medium}$& $\mathtt{Small}$& $\mathtt{Zero}$& $\mathtt{Medium}$& $\mathtt{Large}$ \\
Goal& $\mathtt{Large}$& $\mathtt{Zero}$& $\mathtt{Small}$& $\mathtt{Small}$& $\mathtt{Large}$ \\
Finally blocks in col.& 8& 0& 3& 4& 10 \\
Goal achievement& no& yes& yes& yes& yes \\
\hline\hline
\end{tabular*}
\end{minipage}
\end{table}

\section{Conclusion and future work}
\label{Conclusions}

The combined method for reasoning about actions and robot's beliefs has been presented. The approach is implemented in the Prolog declarative programming language, therefore the method may be tested and modified. The main advantages of the presented approach are: scalability feature, formal language usage, incomplete and uncertain knowledge representation, qualitative reasoning without strictly numerical information. The mentioned features are essential for robotic application which cannot use the quantitative, numerical methods of reasoning. Such situations are common and occur in an unknown environment, when sensors are not able to return precise information or in cases when the reasoning need to be achieved quickly and classical numerical methods need too much time (or too much computational power). The method was implemented for the moving blocks problem, but since more fluents and rules of reasoning might be easily defined, this method can be used for other problems as well.

The future work consists of the reasoning system's extension for further robotic applications, for instance, for the qualitative representation of moving objects in general and qualitative methods of detection. Another idea is to exploit a method for systems where no certain knowledge may be achieved, e.g., when the gain of information always contains a noise that cannot be filtered. The implementation for the real robot and not only simulated one is also taken into consideration. Additionally, the usage of the ExpCog framework \cite{expcog} that has been established in order to integrate different robot reasoning approaches is planned.

\medskip
\section*{Acknowledgments}

The work presented in this paper is supported by the Polish National Science Centre grant 2011/02/A/HS1/00395. The author would like to thank Joanna Goli\'nska -- Pilarek for her valuable criticism and members of the ExpCog project, namely: Mehul Bhatt, Jakob Suchan, Maciej Przybylski for a great chance to work with their ExpCog framework.

\bibliographystyle{splncs}
\bibliography{biblio}

\label{lastpage}
\end{document}